\documentclass[runningheads]{llncs}
\usepackage{graphicx}

\usepackage{tikz}
\usepackage{comment}
\usepackage{amsmath,amssymb} 
\usepackage{color}
\usepackage{algorithm,algcompatible}
\usepackage{algorithm}
\usepackage{algpseudocode}
\usepackage{makecell} 
\usepackage{float}

\usepackage[accsupp]{axessibility}  

\begin{document}
\pagestyle{headings}
\mainmatter
\def\ECCVSubNumber{3665}  

\title{Primitive-based Shape Abstraction \\
via Nonparametric Bayesian Inference} 

\titlerunning{Shape Abstraction via Superquadrics}
%
\author{Yuwei Wu\inst{1} \and
Weixiao Liu\inst{1,2} \and
Sipu Ruan\inst{1} \and
Gregory S. Chirikjian\inst{1*}}
\authorrunning{Y. Wu et al.}
%
\institute{National University of Singapore \and
Johns Hopkins University\\
\email{ \{yw.wu, mpewxl, ruansp, mpegre\}@nus.edu.sg}}

\maketitle

\renewcommand{\thefootnote}{\fnsymbol{footnote}}
\footnotetext[1]{Corresponding author}

\begin{abstract}
3D shape abstraction has drawn great interest over the years. Apart from low-level representations such as meshes and voxels, researchers also seek to semantically abstract complex objects with basic geometric primitives. 
Recent deep learning methods rely heavily on datasets, with limited generality to unseen categories.
Furthermore, abstracting an object accurately yet with a small number of primitives still remains a challenge.
In this paper, we propose a novel non-parametric Bayesian statistical method to infer an abstraction, consisting of an unknown number of geometric primitives, from a point cloud.
We model the generation of points as observations sampled from an infinite mixture of Gaussian Superquadric Taper Models (GSTM).
Our approach formulates the abstraction as a clustering problem, in which: 1) each point is assigned to a cluster via the Chinese Restaurant Process (CRP); 2) a primitive representation is optimized for each cluster, and 3) a merging post-process is incorporated to provide a concise representation.
We conduct extensive experiments on two datasets.
The results indicate that our method outperforms the state-of-the-art in terms of accuracy and is generalizable to various types of objects.

\end{abstract}

\section{Introduction}
Over the years, 3D shape abstraction has received considerable attention. 
Low-level representations such as meshes \cite{deng2020cvxnet,groueix2018papier}, voxels \cite{anwar2006towards,choy20163d}, point clouds \cite{achlioptas2018learning,fan2017point} and implicit surfaces \cite{genova2019learning,mescheder2019occupancy} have succeeded in representing 3D shapes with accuracy and rich features.
However, they cannot reveal the part-level geometric features of an object. 
Humans, on the other hand, are inclined to perceive the environment by parts \cite{pentland1987perceptual}.
Studies have shown that the human visual system makes tremendous use of part-level description to guide the perception of the environment \cite{tversky1984objects}. 
As a result, the part-based abstraction of an object appears to be a promising way to allow a machine to perceive the environment more intelligently and hence perform higher-level tasks like decision-making and planning.
Inspired by those advantages, researchers seek to abstract objects with volumetric primitives, such as polyhedral shapes \cite{roberts1963machine}, spheres \cite{hao2020dualsdf} and cuboids \cite{niu2018im2struct,tulsiani2017learning,yang2021unsupervised,zou20173d}. 
Those primitives, however, are very limited in shape expressivity and suffer from accuracy issues.
Superquadrics, on the other hand, are a family of geometric surfaces that include common shapes such as spheres, cuboids, cylinders, octahedra, and shapes in between, but are only encoded by five parameters. 
By further applying global deformations, they can express shapes such as square frustums, cones, and teardrop shapes. 
Due to their rich shape vocabulary, superquadrics have been widely applied in robotics, \textit{e.g.} grasping \cite{quispe2015exploiting,vezzani2017grasping,vezzani2018improving}, collision detection \cite{ruan2019efficient}, and motion planning \cite{ruan2022closed}.

\begin{figure}
    \centering
    \includegraphics[width=0.9\columnwidth]{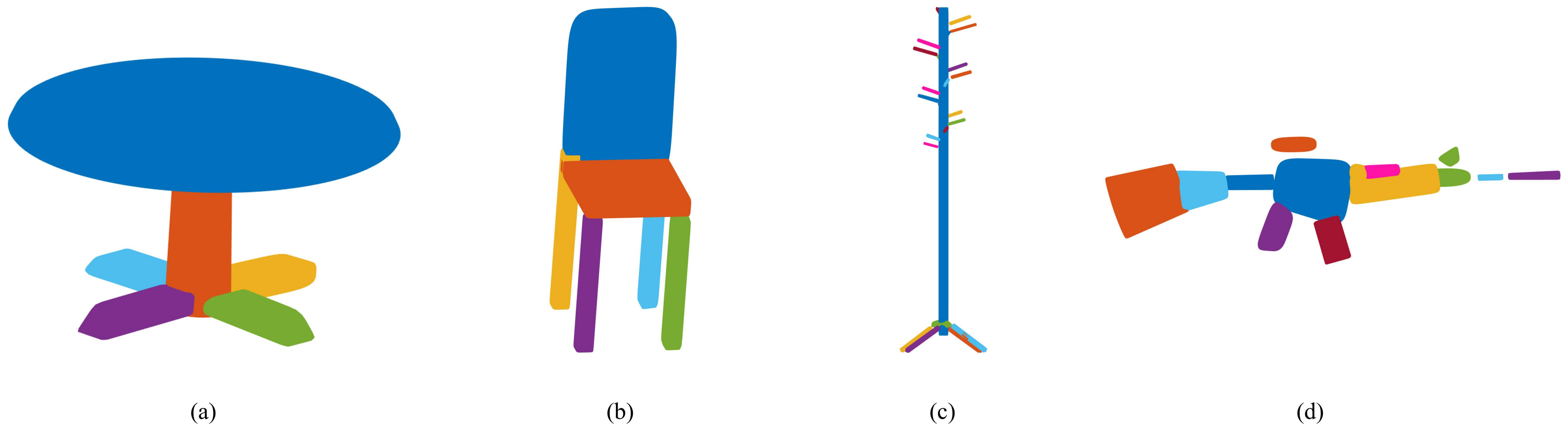}
    \caption{(a)-(d) Examples of multi-tapered-superquadric-based structures of a table, chair, cloth rack, and rifle, inferred by our proposed method.}
    \label{fig:intro}
\end{figure}

The authors of \cite{chevalier2003segmentation,leonardis1997superquadrics} pioneered abstracting superquadric-based representations from complex objects. 
Recently, the authors in \cite{liu2021robust} developed a hierarchical process to abstract superquadric-based structures. 
But, their method necessitates that an object has a hierarchical geometric structure.
In \cite{paschalidou2020learning,paschalidou2019superquadrics}, the authors utilize deep learning techniques to infer superquadric representations from voxels or images.
However, the data-driven approaches show limitations in abstraction accuracy and generality beyond the training dataset.

Our work focuses on accurately abstracting a multi-tapered-superquadric-based representation of a point cloud using a small number of primitives.
By assuming that an object is composed of superquadric-like components, we can regard the problem as a clustering task, which provides a means for us to reason about which portion of the point set can be properly fitted by a single tapered superquadric and thus belongs to the same cluster. 
The collection of tapered superquadrics fitted to each cluster constitutes the multi-tapered-superquadric-based model.
Inspired by the work \cite{liu2021robust} in which the authors construct a Gaussian model around a superquadric, we build a probabilistic model by mixing Gaussian components to account for numerous components of an object.
Since the number of components of an object is unknown in advance and varies case by case, we adapt our model to a nonparametric perspective to assure generality.
Gibbs sampling is applied to infer the posterior distribution, in which we incorporate both an optimization method \cite{liu2021robust} for recovering superquadrics accurately from the point set and a merging process for minimizing the number of primitives, leading to a more exact, compact, and interpretable representation.
Evaluations on Shapenet \cite{chang2015shapenet} and D-FAUST \cite{dfaust:CVPR:2017} corroborate the superior performance of our method in the abstraction of 3D objects.

\section{Related Work}
In this section, we cover the mathematical definition of superquadrics and discuss relevant work on 3D representations.

\subsection{Superquadrics}
Superquadrics \cite{barr1981superquadrics} are a family of geometric surfaces that include common shapes, such as  spheres, cuboids, cylinders, and octahedra, but only encoded by five parameters.
A superquadric surface can be parameterized by $\omega\in(-\pi,\pi]$ and $\eta\in[-\frac{\pi}{2},\frac{\pi}{2}]$:
\begin{equation}
\begin{aligned}
\label{eq:sq}
&\mathbf{p}(\eta, \omega)=\left[\begin{array}{c}
C_{\eta}^{\varepsilon_{1}} \\
a_{z} S_{\eta}^{\varepsilon_{1}}
\end{array}\right] \otimes\left[\begin{array}{l}
a_{x} C_{\omega}^{\varepsilon_{2}} \\
a_{y} S_{\omega}^{\varepsilon_{2}}
\end{array}\right]=\left[\begin{array}{c}
a_{x} C_{\eta}^{\varepsilon_{1}} C_{\omega}^{\varepsilon_{2}} \\
a_{y} C_{\eta}^{\varepsilon_{1}} S_{\omega}^{\varepsilon_{2}} \\
a_{z} S_{\eta}^{\varepsilon_{1}}
\end{array}\right] \\
&C_{\alpha}^{\varepsilon} \triangleq \operatorname{sgn}(\cos (\alpha))|\cos (\alpha)|^{\varepsilon},\, S_{\alpha}^{\varepsilon} \triangleq \operatorname{sgn}(\sin (\alpha))|\sin (\alpha)|^{\varepsilon},
\end{aligned}
\end{equation}
where $\otimes$ denotes the spherical product \cite{barr1981superquadrics}, $\varepsilon_1$ and $\varepsilon_2$ define the sharpness of the shape, and $a_x$, $a_y$, and $a_z$ control the size and aspect ratio. Eq. \ref{eq:sq} is defined within the superquadric canonical frame. 
The expressiveness of a superquadric can be further extended with global deformations such as bending, tapering, and twisting \cite{barr1987global}.
In our work, we apply a linear tapering transformation along z-axis defined as follows:
\begin{equation}
x^\prime= \left(\frac{k_{x}}{a_{3}} z+1\right)x,\,
y^\prime= \left(\frac{k_{y}}{a_{3}} z+1\right)y,\,
z^\prime = z,
\end{equation}
where $-1 \leq k_x,k_y \leq 1$ are tapering factors, $(x,y,z)$ and $(x^\prime,y^\prime,z^\prime)$ are untapered and tapered coordinates, respectively.
To have a superquadric with a general pose, we apply a Euclidean transformation $g=[\mathbf{R}\in SO(3), \mathbf{t}\in \mathbb{R}^3]\in SE(3)$ to it. Thus, a tapered superquadric $\boldsymbol{\mathcal{S}_{\theta}}$ is fully parameterized by $\boldsymbol{\theta}=[\varepsilon_1,\varepsilon_2,a_x,a_y,a_z,g,k_x,k_y]$. 

\begin{figure}[ht]
    \centering
    \includegraphics[width=0.8\columnwidth]{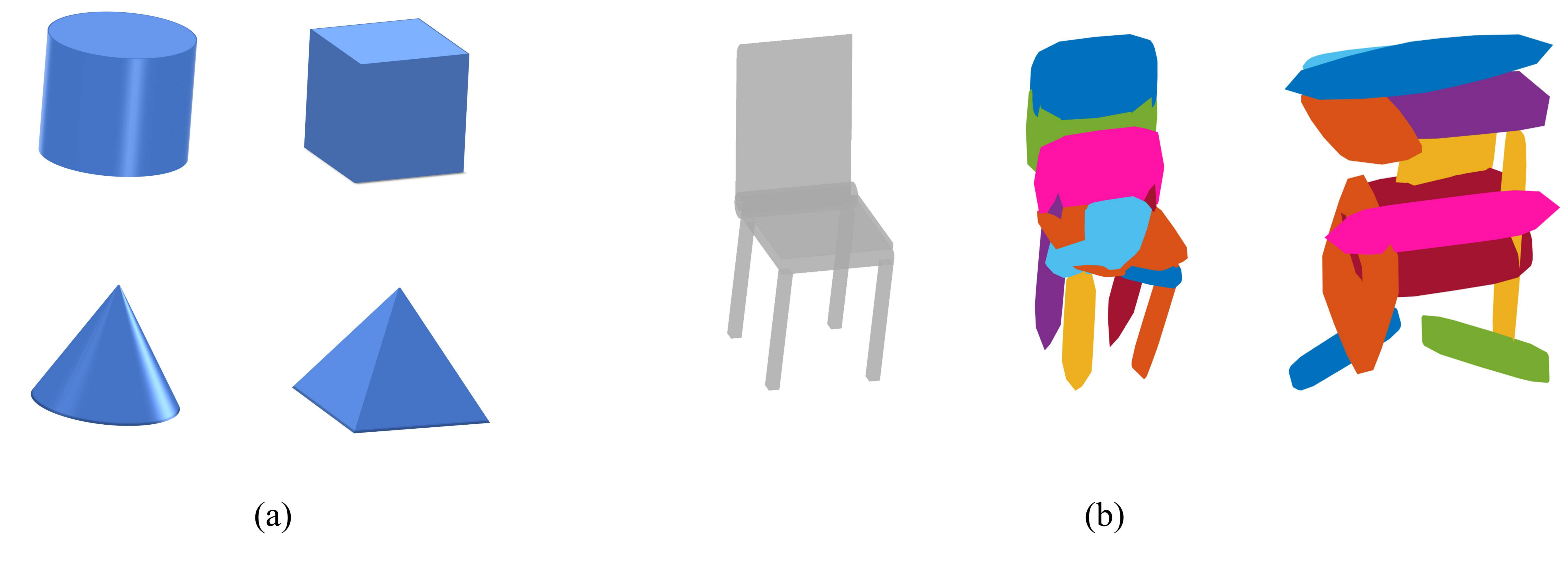}
    \caption{(a) Examples of tapering: a cylinder can be tapered to a cone and a cuboid can be tapered to a square frustum. (b) Part-based models inferred by SQs \cite{paschalidou2019superquadrics}. The left one is the original mesh; the middle one is the superquadrics representation inferred from the network trained on the chair category; the right one is inferred from the network trained on the table category, indicating a limited generality of the DL approach.}
    \label{fig:related work}
\end{figure}

\subsection{3D representations}
Based on how a 3D shape is represented, we can categorize it as a low-level or semantic representation. 

\subsubsection{Low-level Representations}
Standard 3D representations such as voxels, point clouds, and meshes have been extensively studied.
In the work of \cite{anwar2006towards,choy20163d,hane2017hierarchical,riegler2017octnet,slabaugh2004methods,tatarchenko2017octree}, the authors try to recover voxel models from images, which represents the 3D shapes as a regular grid. 
A high-resolution voxel model requires a large amount of memory, which limits its applications. Point clouds are a more memory-efficient way to represent 3D shapes that are utilized in \cite{achlioptas2018learning,fan2017point}, but they fail to reveal surface connectivity. 
Hence, researchers also turn to exploiting meshes \cite{deng2020cvxnet,groueix2018papier,jimenez2016unsupervised,kanazawa2018learning,pan2019deep,wang2018pixel2mesh} to show connections between points. 
Additionally, using implicit surface functions to represent 3D shapes has gained a lot of popularity \cite{atzmon2020sal,chibane2020implicit,genova2020local,genova2019learning,mescheder2019occupancy,park2019deepsdf}. 
Although those representations can capture detailed 3D shapes, they lack interpretability as they cannot identify the semantic structures of objects.

\subsubsection{Part-based Semantic Representations}
To abstract the semantic structures of objects, researchers have attempted to exploit various kinds of volumetric primitives such as polyhedral shapes \cite{roberts1963machine}, spheres \cite{hao2020dualsdf} and cuboids \cite{niu2018im2struct,tulsiani2017learning,zou20173d}. 
However, their results are limited due to the shape-expressiveness of the primitives.
Superquadrics, on the other hand, are more expressive. The authors of \cite{chevalier2003segmentation,leonardis1997superquadrics} are pioneers in abstracting part-based structures from complex objects using superquadrics. They first segmented a complex object into parts and then fitted a single superquadric for each part.
However, their work suffers from limited accuracy.
Recently, the authors in \cite{liu2021robust} proposed a fast, robust, and accurate method to recover a single superquadric from point clouds. 
They exploited the symmetry of the superquadrics to avoid local optima and constructed a probabilistic model to reject outliers. 
Based on the single superquadric recovery, they developed a hierarchical way to represent a complex object with multiple superquadrics. 
The method is effective but requires that an object possess an inherent hierarchical structure.
Another line of primitive-based abstraction is by deep learning \cite{paschalidou2020learning,paschalidou2021neural,paschalidou2019superquadrics}.
Their networks demonstrate the ability to capture fine details of complex objects.
However, the data-driven DL approaches are less generalizable to unseen categories. Besides, they are of a semantic-level approximation, which is lack accuracy.
Instead, our method builds a probabilistic model to reason about primitive-based structures case by case, ensuring generality. Moreover, optimizations are incorporated to yield a more accurate representation for each semantic part.

\section{Method}
\subsection{Nonparametric Clustering Formulation}
In this section, we will show how to cast the problem of superquadric-based abstraction into the nonparametric clustering framework.
To begin with, we model how a random point (observation) $\boldsymbol{x}$ is sampled from a superquadric primitive. 
First, for a superquadric parameterized by $\boldsymbol{\theta}=[\varepsilon_1,\varepsilon_2,a_x,a_y,a_z,g,k_x,k_y]$, a point $\boldsymbol{\mu} \in \mathcal{S}_{\Bar{\boldsymbol{\theta}}}$, where ${\Bar{\boldsymbol{\theta}}}=[\varepsilon_1,\varepsilon_2,a_x,a_y,a_z,(I_3,\mathbf{0}),0,0]$, is randomly selected across the whole surface; 
a noise factor $\tau$ is sampled from an univariate Gaussian distribution $\tau \sim \mathcal{N}(0,\sigma^2)$. 
Then, an point $\Bar{\boldsymbol{x}}$ is generated as 
\begin{equation}
\label{eq:generation of x}
\Bar{\boldsymbol{x}} = (1+ \frac{\tau}{|\boldsymbol{\mu}|})\boldsymbol{\mu},
\end{equation}
where $\tau$ denotes the noise level that shows how far a point deflects the surface. 
After that, we obtain the point $\boldsymbol{x}$ by applying tapering and rigid transformation to $\Bar{\boldsymbol{x}}$, \textit{i.e.} $\boldsymbol{x} = g\circ Taper(\Bar{\boldsymbol{x}})$.
We call the above generative process the \textit{Gaussian Superquadric Taper Model} (GSTM), denoted by:
\begin{equation}
    \boldsymbol{x} \sim GSTM(\boldsymbol{\theta},\sigma^2).
\end{equation}
Subsequently, given a point cloud of an object $\boldsymbol{X} = \{\boldsymbol{x}_i\in \mathbb{R}^{3}|i=1,2,...,N\}$, we assume that each element $\boldsymbol{x}_i$ is generated from some GSTM parameterized by $(\boldsymbol{\theta}_j,\sigma_j^2)$. 
As a result, we consider the point cloud $\boldsymbol{X}$ as sample points generated by a mixture model as follows:
\begin{equation}
    \boldsymbol{X} = \{\boldsymbol{x}_i| \boldsymbol{x}_i \sim \sum_{j = 1}^{K} \omega_j GSTM(\boldsymbol{\theta}_j,\sigma_j^2)\},
\end{equation}
where $\sum_{j=1}^{K} \omega_j = 1$, and each $\omega_j$ denotes the probability that an observation is drawn from $(\boldsymbol{\theta}_j,\sigma_j^2)$. 
Given the observation, we can estimate a set of $\boldsymbol{\theta}$ from the mixture model.
Subsequently, we assume each $\boldsymbol{\theta}_j$ is a shape representation for one semantic part of the object. 
And thus, we attain a set of tapered superquadrics representing the semantic structures for the object. 

The EM algorithm \cite{DEMP1977} is a classical inference to solve a mixture model problem. 
However, EM implementation requires knowledge of $K$ -- the number of components, which in our case is hard to determine beforehand from a raw point cloud. 
Therefore, we handle this difficulty by adapting our model to a nonparametric clustering framework, where we consider $K$ to be infinite.
\begin{figure}
    \centering
    \includegraphics[width=0.4\columnwidth]{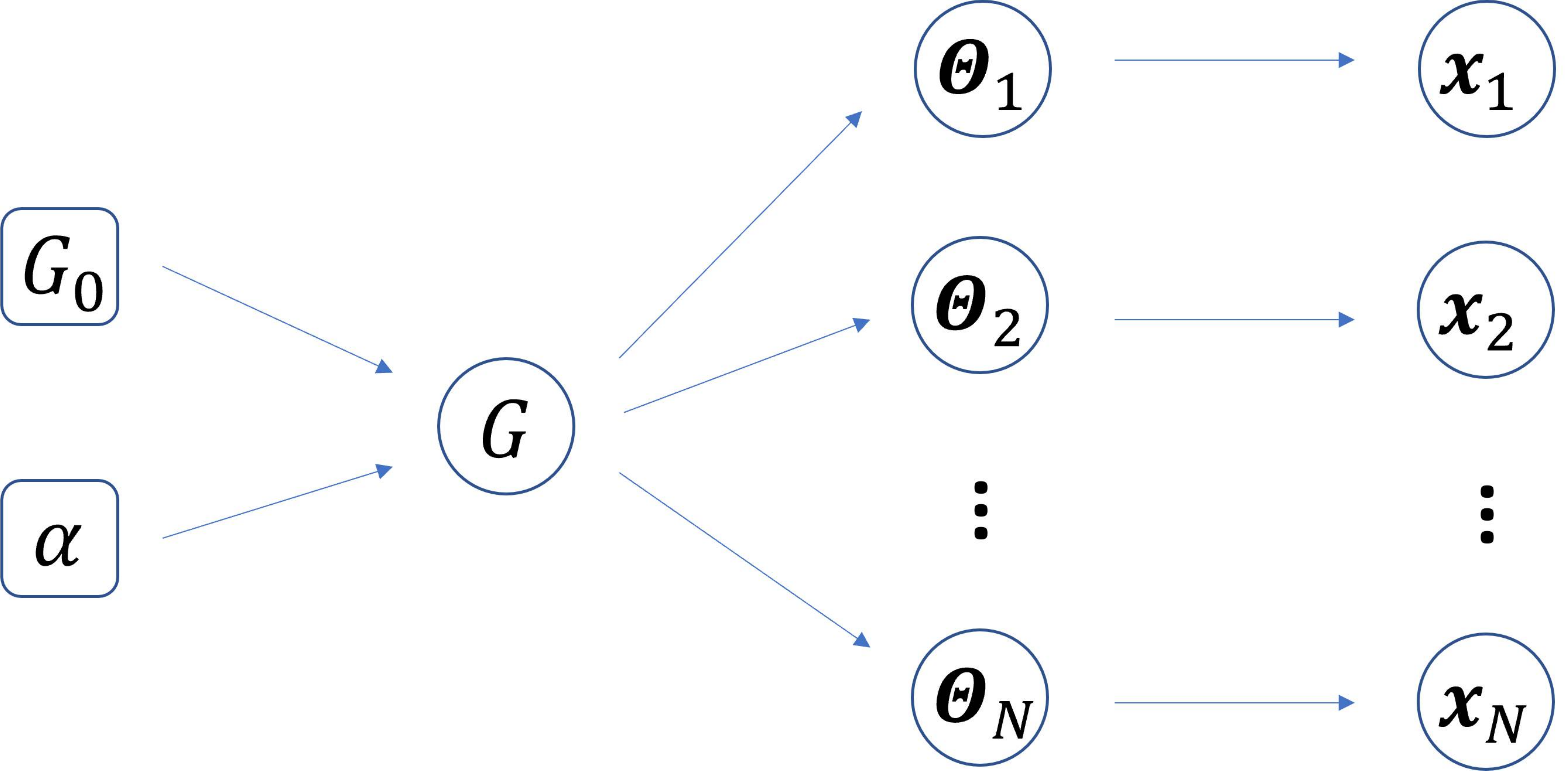}
    \caption{Generative process of a point cloud. Notice that $\boldsymbol{x}_i$ and $\boldsymbol{x}_j$ where $i\neq j$ could be sampled from the same $\boldsymbol{\Theta}$. Since we only draw finite samples, the number of cluster is in fact finite.}
    \label{fig:DP}
\end{figure}

To deal with the mixture model with infinitely many components, we introduce the \textit{Dirichlet Process} (DP) into our formulation.
A DP, parameterized by a base distribution $G_0$ and a concentration factor $\alpha$, is a distribution over distributions:
\begin{equation}
    G \sim DP(G_0,\alpha), 
\end{equation}
which is equivalent to
\begin{equation}\\ G = \sum_{j=1}^{\infty} \omega_j \delta _{\boldsymbol{\Theta}_j}, \
    \boldsymbol{\Theta}_j \sim G_0, \ \pi \sim GEM(\alpha),
\end{equation}
where $\boldsymbol{\Theta}_j = (\boldsymbol{\theta}_j, \sigma_j^2)$, sampled from $G_0$, is the parameter of the $j$th GSTM, $\delta$ is the indicator function which evaluates to zero everywhere, except for $\delta _{\boldsymbol{\Theta}_j}(\boldsymbol{\Theta}_j) = 1$,  
$\pi=(\omega_1,\omega_2,...)$, $\sum_{i=1}^\infty \omega_i=1$, and GEM is the Griffiths, Engen and McCloskey distribution \cite{gnedin2001characterization}.
Therefore, an observation $\mathbf{x}_i$ is regarded as sampled from:
\begin{equation}
    \label{eq:generative model}
    z_i\sim \pi, \boldsymbol{x}_i \sim GSTM(\boldsymbol{\Theta}_{z_i}),
\end{equation}
where $z_i$ is a latent variable sampled from a categorical distribution
parameterized by $\pi$, indicating the membership of $\boldsymbol{x}_i$.
Fig. \ref{fig:DP} illustrates the process. 

Even though we have an infinite mixture model, in practice we only draw finite samples, which means the number of clusters is actually finite. 
One advantage of our formulation is that we do not need to impose any constraints on $K$, which is inferred from the observation $\boldsymbol{X}$. 
On the other hand, unlike learning-based approaches that require a large amount of training data, our method reasons about primitive-based structures case by case, relying entirely on the geometric shapes of the object.
These two facts contribute to increasing the generality of being able to cope with objects of varying shapes and component counts.

\subsection{Optimization-based Gibbs Sampling}
We apply Bayesian inference to solve the mixture model problem, where the goal is to infer the posterior distribution of the parameters $\Tilde{\boldsymbol{\theta}}=\{\boldsymbol{\theta}_1,\boldsymbol{\theta}_2,...,\boldsymbol{\theta}_K\}, \Tilde{\sigma}^2=\{{\sigma}_1^2,{\sigma}_2^2,...,{\sigma}_K^2\}$ and the latent variables $\boldsymbol{Z}=\{z_1,z_2,...,z_N\}$ given the observation $\mathbf{X}$:
\begin{equation}
    \label{eq:posteriors}
    p(\Tilde{\boldsymbol{\theta}}, \Tilde{\sigma}^2,\boldsymbol{Z} \mid  \boldsymbol{X}).
\end{equation}

\begin{algorithm}
\caption{Optimization-based Gibbs sampling}\label{alg:gibbs}
\begin{algorithmic}
    \State \textbf{Input:} $\boldsymbol{X}=\{\boldsymbol{x}_1,\boldsymbol{x}_2,...,\boldsymbol{x}_N\}$
    \State \textbf{Output:} $ \{\Tilde{{\boldsymbol{\theta}}}^t, \Tilde{\sigma}^{2^t}, \boldsymbol{Z}^t\}_{t=1}^T$
    
    \State \textbf{initialize} $\{\Tilde{\boldsymbol{\theta}}^t, \Tilde{\sigma}^{2^t}, \boldsymbol{Z}^t\}$ for $t = 0$ by K-means clustering
    \For{$t=1,2,...,T$}
        \State 1. draw a sample $\boldsymbol{Z}^\prime $ for $\boldsymbol{Z}$, where  $\boldsymbol{Z}^\prime  \sim p(\boldsymbol{Z} \mid \boldsymbol{X},\Tilde{\boldsymbol{\theta}}^t, \Tilde{\sigma}^{2^t})$
        
        \State 2. optimize each element $\boldsymbol{\theta}_j$ of $\Tilde{\boldsymbol{\theta}}$ conditioned on $\{\boldsymbol{Z}^\prime, \boldsymbol{X}, \Tilde{\sigma}^{2^t} \}$, and let $\Tilde{\boldsymbol{\theta}}^\prime$ be the optimized $\Tilde{\boldsymbol{\theta}}$
        
        \State 3. draw a sample $\Tilde{\sigma}^{2^\prime}$ for $\Tilde{\sigma}^2$, where $\Tilde{\sigma}^{2^\prime} \sim
        p(\Tilde{\sigma}^2 \mid \boldsymbol{X}, \boldsymbol{Z}^\prime, \Tilde{\boldsymbol{\theta}}^\prime)$
        
        \State 4. let  $\{\Tilde{\boldsymbol{\theta}}^{t+1}, \Tilde{\sigma}^{2^{t+1}}, \boldsymbol{Z}^{t+1}\} = \{\Tilde{\boldsymbol{\theta}}^\prime, \Tilde{\sigma}^{2^\prime},\boldsymbol{Z}^\prime \}$
        
    \EndFor
\end{algorithmic}
\end{algorithm}
However, in practice, the Eq. \ref{eq:posteriors} is intractable to obtain in a closed-form.
Thus, we apply Gibbs sampling \cite{gelfand2000gibbs}, an approach to estimating the desired probability distribution via sampling.
Apart from sampling, we also incorporate an optimization process, which is used to obtain an accurate superquadric representation for each cluster. 
The following algorithm \ref{alg:gibbs} shows how optimization-based Gibbs sampling works in our case. 
In the following sections, we will derive and demonstrate explicitly how to obtain each parameter.

\subsubsection{Sample $\boldsymbol{Z}$}
\begin{figure}[H]
    \centering
    \includegraphics[width=0.2\columnwidth]{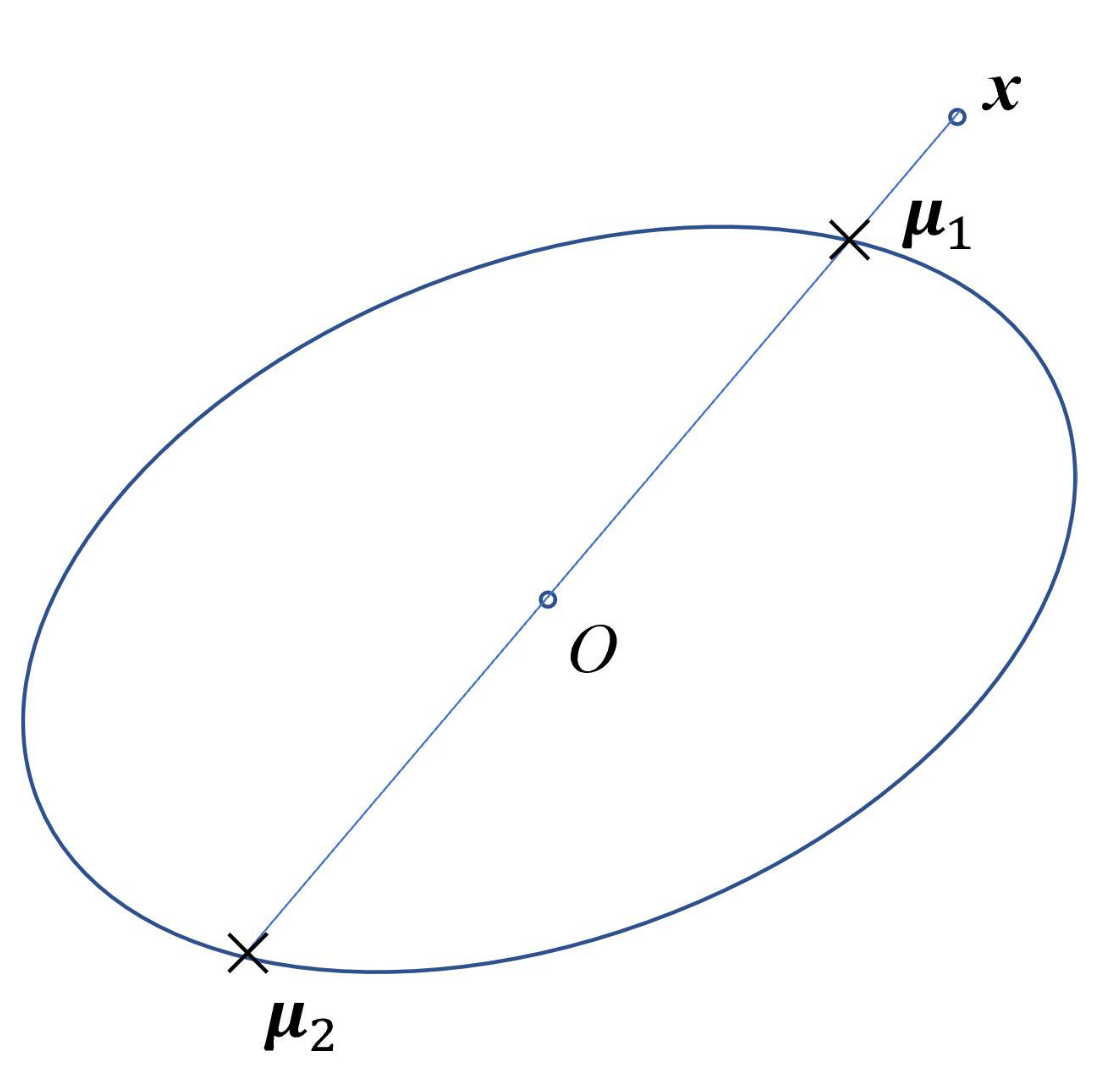}
    \caption{Demonstration for computing sampling density of $\boldsymbol{x}$. According to GSTM, $\boldsymbol{\mu}_1$ and $\boldsymbol{\mu}_2$ are the only two points accounting for the generation of $\boldsymbol{x}$.}
    \label{fig:sampleZ}
\end{figure}

To begin with, as defined in Eq. \ref{eq:generation of x}, we have the sampling distribution of $\boldsymbol{x}$
\begin{equation}
    \label{eq:prob of x}
    \begin{split}
        p(\boldsymbol{x}\mid \boldsymbol{\theta},\sigma^2) = \frac{1}{2\sqrt{2\pi}\sigma}\exp\left(-\frac{\| \boldsymbol{x}-\boldsymbol{\mu}_1(\boldsymbol{\theta},\boldsymbol{x})\|_2^2}{2\sigma^2}\right) + \\
        \frac{1}{2\sqrt{2\pi}\sigma}\exp\left(-\frac{\| \boldsymbol{x}-\boldsymbol{\mu}_2(\boldsymbol{\theta},\boldsymbol{x})\|_2^2}{2\sigma^2}\right),
    \end{split}
\end{equation}
where $\boldsymbol{\mu}_1$ and $\boldsymbol{\mu}_2$ are two intersection points between the superquadric surface and the line joining the superqadric's origin and $\boldsymbol{x}$, as Fig. \ref{fig:sampleZ} shows.
We denote $\boldsymbol{\mu}_1$ as the closer intersection point to $\boldsymbol{x}$. In the general case, Eq. \ref{eq:prob of x} is dominated by the former part, and hence we let
\begin{equation}
    \label{eq:appro of x}
     p(\boldsymbol{x}\mid \boldsymbol{\theta},\sigma^2) \approx \frac{1}{2\sqrt{2\pi}\sigma}\exp\left(-\frac{\| \boldsymbol{x}-\boldsymbol{\mu}_1(\boldsymbol{\theta},\boldsymbol{x})\|_2^2}{2\sigma^2}\right).
\end{equation}
By further examination, we discover that the term $\|\boldsymbol{x}-\boldsymbol{\mu}_1(\boldsymbol{\theta},\boldsymbol{x})\|_2$ is the radial distance between a point and a superquadric as defined in \cite{gross1988error}.
We denote it by $d(\boldsymbol{\theta},\boldsymbol{x})$. 
Integrating out the $\boldsymbol{\theta}$ and $\sigma^2$ gives:
\begin{equation}
    \label{eq:p(x)}
    p(\boldsymbol{x}) = \int_{\boldsymbol{\theta},\sigma^2}p(\boldsymbol{x}\mid \boldsymbol{\theta},\sigma^2) p(\boldsymbol{\theta},\sigma^2) \ d\boldsymbol{\theta} d\sigma^2 \approx 
    \int_{\boldsymbol{\theta},\sigma^2} \frac{p(\boldsymbol{\theta},\sigma^2)}{2\sqrt{2\pi}\sigma} \exp\left(-\frac{d^2(\boldsymbol{\theta},\boldsymbol{x})}{2\sigma^2}\right) \ d\boldsymbol{\theta} d\sigma^2.
\end{equation}
Eq. \ref{eq:p(x)} denotes the prior predictive density of $\boldsymbol{x}$ and is intractable to compute in closed form. It can be approximated by Monte Carlo sampling or approximated by a constant \cite{hayden2020nonparametric}.

In our work, we treat $p(\boldsymbol{x})$ as a tunable hyper-parameter and denote it as $p_0$.
To sample membership for each point $\boldsymbol{x}_i$, we have 
\begin{equation}
\begin{split}
    \label{eq:existing cluster}
    p\left(z_{i}=j \mid \boldsymbol{Z}_{-i}, \boldsymbol{\theta}_j, \sigma_{j}^2, \boldsymbol{X}, \alpha\right) \propto p\left(z_{i}=j \mid \boldsymbol{Z}_{-i}, \alpha\right) p\left(\boldsymbol{x}_{i} \mid \boldsymbol{\theta}_j, \sigma_{j}^2, \boldsymbol{Z}_{-i}\right) \\
    \propto \frac{n_{-i, j}}{N-1+\alpha} p(\boldsymbol{x}_i\mid \boldsymbol{\theta}_j,\sigma_j^2) =  \frac{n_{-i, j}}{N-1+\alpha} \frac{1}{2\sqrt{2\pi}\sigma_j}\exp\left(-\frac{d^2(\boldsymbol{\theta}_j,\boldsymbol{x}_i)}{2\sigma_j^2}\right),
\end{split}
\end{equation}
and
\begin{equation}
\begin{split}
    \label{eq:new cluster}
    p\left(z_{i}=K+1 \mid \boldsymbol{Z}_{-i}, \boldsymbol{\theta}_j, \sigma_{j}^2, \boldsymbol{X}, \alpha\right) \propto p\left(z_{i}=K+1 \mid \alpha\right) p\left(\boldsymbol{x}_{i} \mid \boldsymbol{\theta}_j, \sigma_{j}^2, \boldsymbol{Z}_{-i}\right) \\
      \propto \frac{\alpha}{N-1+\alpha} p(\boldsymbol{x}_i) =  \frac{\alpha}{N-1+\alpha} p_0,
\end{split}
\end{equation}
where $\alpha$ is the concentration factor of DP, $\boldsymbol{Z}_{-i}$ denotes $\boldsymbol{Z}$ excluding $z_i$, and $n_{-i, j}$ is the number of points belonging to cluster $j$, excluding $\boldsymbol{x}_i$. Eq. \ref{eq:existing cluster} computes the probability that $\boldsymbol{x}_i$ belongs to some existing cluster, whereas Eq. \ref{eq:new cluster} determines the probability of generating a new cluster. 
The term $p\left(z_{i}=j \mid \boldsymbol{Z}_{-i}, \alpha\right)$ of Eq. \ref{eq:existing cluster} and $p\left(z_{i}=K+1 \mid \alpha\right)$ of Eq. \ref{eq:new cluster} come from the \textit{Chinese Restaurant Process} (CRP), where a point tends to be attracted by a larger population and has a fixed probability to generate a new group. 
The term $p\left(\boldsymbol{x}_{i} \mid \boldsymbol{\theta}_j, \sigma_{j}^2, \boldsymbol{Z}_{-i}\right)$ reasons about what the likelihood is that $\boldsymbol{x}_i$ belongs to some existing cluster or a new one, based on the current $\Tilde{\boldsymbol{\theta}}=\{\boldsymbol{\theta}_1,\boldsymbol{\theta}_2,...,\boldsymbol{\theta}_K\} \text{ and } \Tilde{\sigma}^2=\{{\sigma}_1^2,{\sigma}_2^2,...,{\sigma}_K^2\}$. 
After the assignment of all points, some existing clusters may be assigned with none of the points and we remove those empty clusters. 
Thus, the $K$ keeps changing during each iteration. To increase the performance, we incorporate a splitting process before sampling $\boldsymbol{Z}$. Details are presented in the supplementary.

\subsubsection{Optimize $\Tilde{\boldsymbol{\theta}}$}
By independence between individual $\boldsymbol{\theta}$, the density function of each $\boldsymbol{\theta}_j$ is conditioned only on $\boldsymbol{X}^j \text{ and } \sigma_j^2$ as follows:
\begin{equation}
\begin{split}
    p(\boldsymbol{\theta}_j \mid \boldsymbol{X}^j, \sigma_j^2),
\end{split}
\end{equation}
where $\boldsymbol{X}^j=\{\boldsymbol{x}_l \mid \boldsymbol{x}_l \in \boldsymbol{X},z_l = j\}$, \textit{i.e.} the set of points belonging to cluster $j$. By assuming that the prior for $\boldsymbol{\theta}$ is an uniform distribution, we have 
\begin{equation}
    \label{eq:theta1}
    p(\boldsymbol{\theta}_j \mid \boldsymbol{X}^j, \sigma_j^2) \propto p(\boldsymbol{\theta}_j) p(\boldsymbol{X}^j \mid \boldsymbol{\theta}_j, \sigma_j^2) \propto p(\boldsymbol{X}^j \mid \boldsymbol{\theta}_j, \sigma_j^2),
\end{equation}
where
\begin{equation}
    \label{eq:theta2}
    p(\boldsymbol{X}^j \mid \boldsymbol{\theta}_j, \sigma_j^2) = \prod_l \frac{1}{2\sqrt{2\pi}\sigma_j} \exp\left(-\frac{d^2(\boldsymbol{\theta}_j,\boldsymbol{x}_l)}{2\sigma_j^2}\right).
\end{equation}
Combining Eq. \ref{eq:theta1} and Eq. \ref{eq:theta2} gives
\begin{equation}
      \label{eq:theta3}
      p(\boldsymbol{\theta}_j \mid \boldsymbol{X}^j, \sigma_j^2) \propto \prod_l \exp \left(-\frac{d^2(\boldsymbol{\theta}_j,\boldsymbol{x}_l)}{2\sigma_j^2}\right) = 
      \exp \left(-\sum_l \frac{d^2(\boldsymbol{\theta}_j,\boldsymbol{x}_l)}{2\sigma_j^2}      \right).
\end{equation}
Gibbs sampling requires sampling $\boldsymbol{\theta}_j$ from Eq. \ref{eq:theta3}. However, directly sampling from Eq. \ref{eq:theta3} is difficult due to its complexity.
Instead, optimization is used as a substitute for the sampling process, which, we believe, is a reasonable replacement.
By inspecting Eq. \ref{eq:theta3} more closely, we discover that the $\boldsymbol{\theta}_{j}$ minimizing $\sum d^2(\boldsymbol{\theta}_j,\boldsymbol{x}_l)$ maximizes the density function.
Therefore, an optimized $\boldsymbol{\theta}_{j}$ has relatively higher likelihood to be close to the actual sample of $\boldsymbol{\theta}_{j}$ drawn from $p(\boldsymbol{\theta}_j \mid \boldsymbol{X}^j, \sigma_j^2)$. 
And closeness implies similar shapes. 
Additionally, we recognize that optimizing Eq. \ref{eq:theta3} can be regarded as a single superquadric recovery problem, which requires the abstraction of an optimal superquadric primitive from the cluster points $\boldsymbol{X}^j$. 
In other words, we fit an optimal superquadric to each cluster, and those superquadrics will affect the membership of each point in the subsequent iteration, which is a process similar to EM. 
As a result, we use the robust and accurate recovery algorithm \cite{liu2021robust}, which yields an optimal superquadric with high fidelity, to acquire each $\boldsymbol{\theta}_j$ in replacement of the sampling.

\subsubsection{Sample $\Tilde{\sigma}^2$}
Similarly, by independence, we have 
\begin{equation}
    \begin{split}
        \sigma_j^{2^\prime} \sim p(\sigma_j^2 \mid \boldsymbol{X}^j, \boldsymbol{\theta}_j) \\
        \Tilde{\sigma}^{2^\prime}=\{{\sigma}_1^{2^\prime},{\sigma}_2^{2^\prime},...,{\sigma}_K^{2^\prime}\}.
    \end{split}
\end{equation}
We also assume the non-informative prior for $\sigma^2$ is the uniform distribution, which gives
\begin{equation}
    \label{eq:sigma2}
    p(\sigma_j^2 \mid \boldsymbol{X}^j,\boldsymbol{\theta}_j) \propto p(\sigma_j^2)p(\boldsymbol{X}^j \mid \boldsymbol{\theta}_j,\sigma_j^2) \propto p(\boldsymbol{X}^j \mid \boldsymbol{\theta}_j,\sigma_j^2).
\end{equation}
Combining Eq. \ref{eq:theta2} and Eq. \ref{eq:sigma2} gives
\begin{equation}
    \begin{split}
    p(\sigma_j^2 \mid \boldsymbol{X}^j,\boldsymbol{\theta}_j) \propto \prod_l \frac{1}{2\sqrt{2\pi}\sigma_j} \exp\left(-\frac{d^2(\boldsymbol{\theta}_j,\boldsymbol{x}_l)}{2\sigma_j^2}\right) \\
    \propto \left( \frac{1}{\sigma_j} \right)^{n_j}\exp \left(-\sum_l \frac{d^2(\boldsymbol{\theta}_j,\boldsymbol{x}_l)}{2\sigma_j^2}\right),
    \end{split}
\end{equation}
where $n_j$ is the number of $\boldsymbol{X^j}$. Let $D = \sum_l d^2(\boldsymbol{\theta}_j,\boldsymbol{x}_l)$ and $\gamma_j = \frac{1}{\sigma_j^2}$. By change of variable, we have 
\begin{equation}
    \gamma_j^\prime \sim p(\gamma_j \mid \boldsymbol{X}^j,\boldsymbol{\theta}_j) \propto \gamma_j^{\frac{n_j-3}{2}}\exp(-\frac{D}{2}\gamma_j).
\end{equation}
Hence, $\gamma_j$ follows a gamma distribution with shape parameter $\frac{n_j-1}{2}$ and scale parameter $\frac{2}{D}$. In other words, 
\begin{equation}
\begin{aligned}
    &\sigma_j^{2^\prime} = \frac{1}{\gamma_j^\prime}\\
    &\gamma_j^\prime \sim \boldsymbol{\Gamma}\left(\frac{n_j-1}{2},\frac{2}{D}\right).
\end{aligned}
\end{equation}
$D$ reflects how good the optimized superquadric fits the cluster points. With lower value of $D$, $\gamma_j^\prime$ will have a better chance to be higher and hence $\sigma_j^{2^\prime}$ will be smaller. In other words, the better the fitting is, the smaller the noise level will be.

\subsection{Merging process}
We observe that our method yields structures consisting of excessive components, resulting in less interpretability. 
Therefore, we design a merging post-process minimizing component numbers while maintaining accuracy. 
Specifically, for any two clusters represented by two superquadrics, we make a union of the two sets of points into one set, from which we recover a superquadric.
If the newly recovered superquadric turns out to be a good fit for the new point set, we will merge these two clusters into one, and replace the two original superquadrics with the newly fitted one. 
Detailed formulations and procedures are presented in the supplementary.  

\section{Experiment}
In this section, we demonstrate our approach to abstracting part-level structures exhibits high accuracy, compared with state-of-the-art part-based abstraction method \cite{paschalidou2019superquadrics}. We do not compare with the work of \cite{paschalidou2020learning,paschalidou2021neural} since their work focuses mainly on abstracting 3D shapes from 2D images.
We also include a simple clustering method as a baseline, where the point cloud is parsed into $K$ clusters via K-means and each cluster is then represented by an optimized superquadric \cite{liu2021robust}. 
We conduct experiments on the ShapeNet dataset \cite{chang2015shapenet} and the D-FAUST dataset \cite{dfaust:CVPR:2017}.
The ShapeNet is a collection of CAD models of various common objects such as tables, chairs, bottles, etc. 
On the other hand, the D-FAUST dataset contains point clouds of 129 sequences of 10 humans performing various movements, \textit{e.g.}, punching, shaking arms, and running. Following \cite{paschalidou2019superquadrics}, we evaluate the results with two metrics, Chamfer $L_1\text{-distance}$ and Intersection over Union (IoU). Detailed computations of the two metrics are discussed in the supplementary.

\textbf{Initialization:} we parse the point cloud into $K$ components based on the K-means clustering algorithm. Although we specify the value of $K$ initially, the final value of $K$ will be inferred by our nonparametric model and vary from case to case.
The latent variable $z_i$ of each point $\boldsymbol{x}_i$ is assigned accordingly.
Subsequently, each cluster is represented by an ellipsoid $\boldsymbol{\theta}_j^0$ whose moment-of-inertial (MoI) is four times smaller than the MoI of the cluster points; each $\sigma_j^{2^0}$ is randomly sampled within $(0,1]$. 
We set the number of the sampling iteration to be $T=30$, concentration factor $\alpha=0.5$, and $p_0=0.1$. 

\begin{figure}[t]
    \centering 
    \includegraphics[width=1\columnwidth]{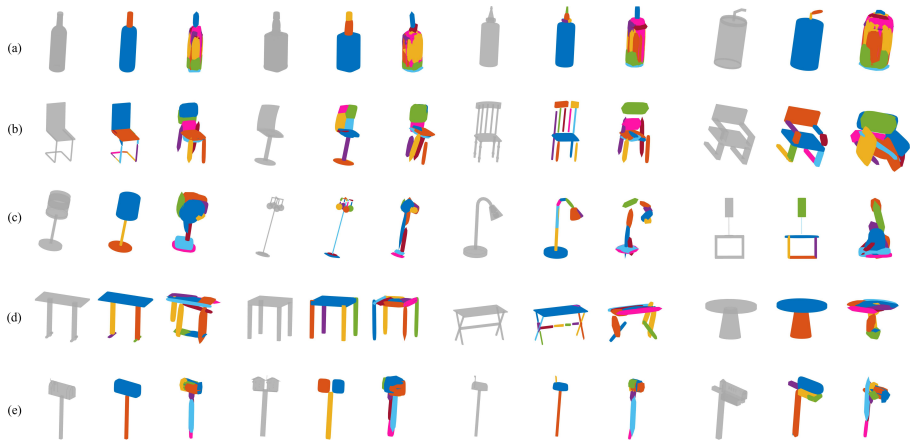}
    \caption{Qualitative results of 3D abstraction on various objects. The left ones are the original meshes, the middle ones are our inferred results, and the right ones are inferred from SQs \cite{paschalidou2019superquadrics}. (a) Bottle, (b) chair, (c) lamp, (d) table, and (e) mailbox.}
    \label{fig:shapenet}
\end{figure}
\subsection{Evaluation on ShapeNet}
We choose seven different types of objects among all of the categories.
For deep learning training, we randomly divide the data of each object into two sets -- a training set (80\%) and a testing set (20\%), and we compare the results on the testing set.
Since ShapeNet only provides meshes, we first densely sample points on the mesh surfaces and then downsample the point clouds to be around 3500 points. 
For all categories, we set the $K=30$. The result is summarized in Table \ref{Tab:shapenet}, where w/om denotes our method excluding the merging process. 
Our method outperforms the state-of-the-art \cite{paschalidou2019superquadrics} and the K-means baseline significantly on all object types. 
Excluding the merging process improves accuracy but increases the number of primitives, making the abstracted models less interpretable. 
Therefore, we believe merging is beneficial and important because it reduces the primitive numbers while maintaining excellent accuracy, which improves interpretability.
Unlike the learning-based method, which is a semantic-level approximation, our method infers the part-based representation in an optimization-based manner. As a result, our method yields a more geometrically accurate primitive-based structure, yet with a compact number of primitives. 
A qualitative comparison between our method and SQs \cite{paschalidou2019superquadrics} is depicted in Fig. \ref{fig:shapenet}.

\begin{figure}[ht]
    \centering
    \includegraphics[width=0.9\columnwidth]{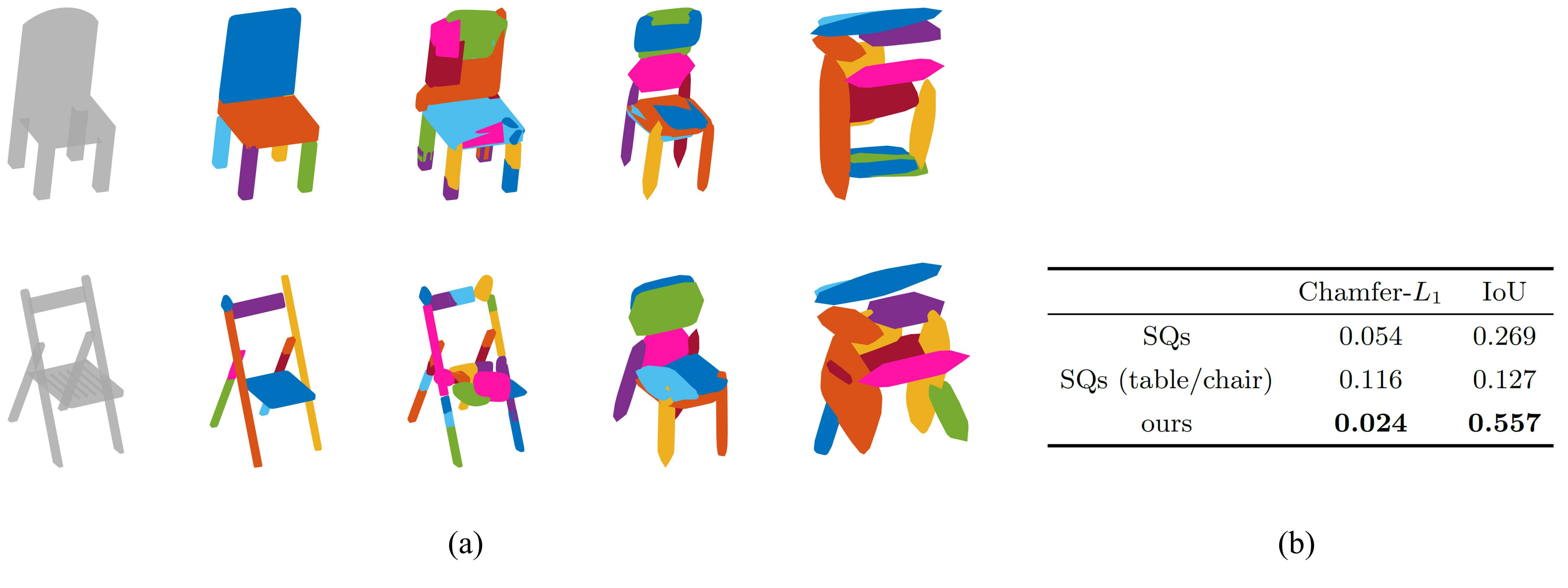}
    \caption{(a) Comparison between different results inferred by different models. From left to right: the original meshes, results inferred by our method, results inferred by our method excluding merging, results inferred by the baseline trained on the chair category, and results inferred by the baseline trained on the table category, (b) quantitative results showing that the baseline method has limited generality. The (table/chair) means that a network trained on the table category is used to predict the chair category.}
    \label{fig:shapenet2}
\end{figure}

\begin{table}[H]
    \setlength{\tabcolsep}{4pt} 
    \renewcommand{\arraystretch}{1.2} 
    \centering
    \caption{Quantitative results on ShapeNet}
    \begin{tabular}{c||cccc||cccc} 
    \Xhline{1.0pt} 
    & \multicolumn{4}{c||}{ Chamfer-$L_{1}$} & \multicolumn{4}{c}{ IoU} \\
    Category & K-means & SQs \cite{paschalidou2019superquadrics}  & w/om & Ours & K-means & SQs  & w/om & Ours\\
    \hline 
    bottle  & 0.064 & 0.037 & 0.026 & $\boldsymbol{0.019}$ & 0.552 & 0.596 & 0.618 & $\boldsymbol{0.656}$ \\
    can     & 0.086 & 0.032 & 0.028 & $\boldsymbol{0.014}$ & 0.690 & 0.736 & $\boldsymbol{0.803}$ & 0.802 \\
    chair   & 0.065 & 0.054 & $\boldsymbol{0.018}$ & 0.024 & 0.433 & 0.269 & $\boldsymbol{0.577}$ & 0.557\\
    lamp    & 0.066 & 0.065 & $\boldsymbol{0.020}$ & 0.024 & 0.354 & 0.190 & $\boldsymbol{0.425}$ & 0.414\\
    mailbox & 0.054 & 0.059 & 0.019 & $\boldsymbol{0.019}$ & 0.529 & 0.400 & $\boldsymbol{0.687}$ & 0.686\\
    rifle   & 0.018 & 0.022 & $\boldsymbol{0.009}$ & 0.013 & 0.517 & 0.291 & $\boldsymbol{0.594}$ & 0.536\\
    table   & 0.060 & 0.057 & $\boldsymbol{0.018}$ & 0.021 & 0.374 & 0.194 & $\boldsymbol{0.536}$ & 0.512\\
    \hline 
    mean    & 0.057 & 0.053 & $\boldsymbol{0.017}$ & 0.021 & 0.410 & 0.242 & $\boldsymbol{0.547}$ & 0.526\\
    \Xhline{1.0pt} 
    \end{tabular}
    
    \label{Tab:shapenet}
\end{table}

\begin{table}[H]
    \setlength{\tabcolsep}{4pt} 
    \renewcommand{\arraystretch}{1.2} 
    \centering
    \begin{tabular}{c||cccc} 
    \Xhline{1.0pt} 
    & \multicolumn{4}{c}{\#primitives} \\
    Category & K-means & SQs & w/om & Ours\\
    \hline 
    bottle  & 30 & 8  & 7.4 & 6.8 \\
    can     & 30 & 7  & 13.6 & 1.1  \\
    chair   & 30 & 10 & 26.9 & 13.6  \\
    lamp    & 30 & 10 & 24.8 & 9.5 \\
    mailbox & 30 & 10 & 18.8 & 3.1 \\
    rifle   & 30 & 7  & 20.0 & 7.6 \\
    table   & 30 & 11 & 25.1 & 9.5 \\
    \hline 
    \Xhline{1.0pt} 
    \end{tabular}
    \label{Tab:shapenet_continued}
\end{table}
Furthermore, generality is noteworthy. To attain the reported accuracy, the baseline method needs to be trained on a dataset of a specified item category, respectively. 
A network trained on one item is difficult to generalize to another as Fig. \ref{fig:shapenet2} shows. 
In contrast, our probabilistic formulation reasons about the part-based representation case by case, and the nonparametric formulation makes it possible to adapt to various shapes with varying component numbers.

\subsection{Evaluation on D-FAUST}
\begin{figure}[ht]
    \centering
    \includegraphics[width=1\columnwidth]{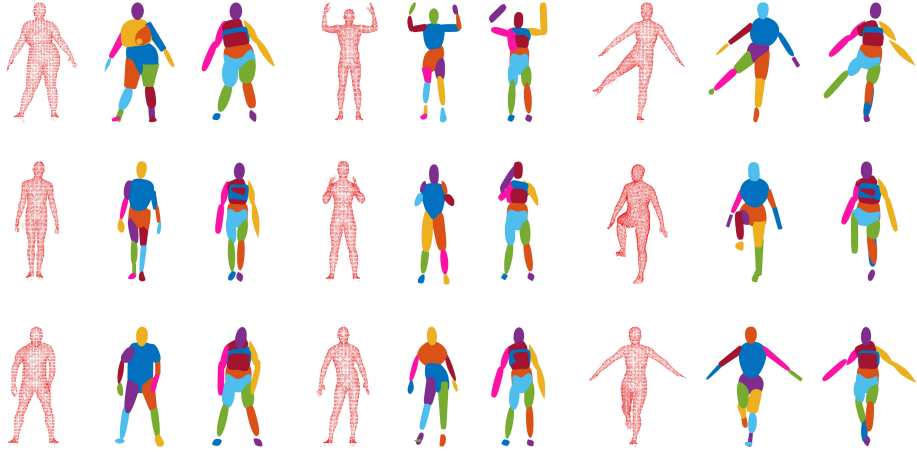}
    \caption{Abstraction results on D-FAUST dataset. The left ones are the original point clouds, the middle ones are inferred by our method, and the right ones are from SQs \cite{paschalidou2019superquadrics}. }
    \label{fig:dfaust}
\end{figure}

We follow the same split strategy in \cite{paschalidou2020learning} and divide the dataset into training (91), testing (29), and validation (9). 
Likewise, we compare results on the testing set. 
For our method, we downsample the point clouds to be around 5500 points and set $K$ to be 30, as well. 
The results are shown in table \ref{tab:dfaust}. Fig. \ref{fig:dfaust} illustrates examples of inferred representations. We can observe that our model can accurately capture the major parts of humans, \textit{i.e.} heads, chests, arms, forearms, hips, thighs, legs, and feet, even when they are performing different movements.

\begin{table}
    \setlength{\tabcolsep}{4pt} 
    \renewcommand{\arraystretch}{1.2} 
    \centering
    \caption{Quantitative results on D-FAUST}
    \begin{tabular}{ccc}
    \Xhline{1.0pt} 
         &  Chamfer-$L_{1}$ & IoU\\
         \hline
        SQs\cite{paschalidou2019superquadrics} & 0.0473                & 0.7138 \\
        ours       & $\boldsymbol{0.0335}$ & $\boldsymbol{0.7709}$ \\
    \Xhline{1.0pt} 
    \end{tabular}
    \label{tab:dfaust}
\end{table}

\subsection{Extension: Point Cloud Segmentation}
\begin{figure}
    \centering
    \includegraphics[width=0.8\columnwidth]{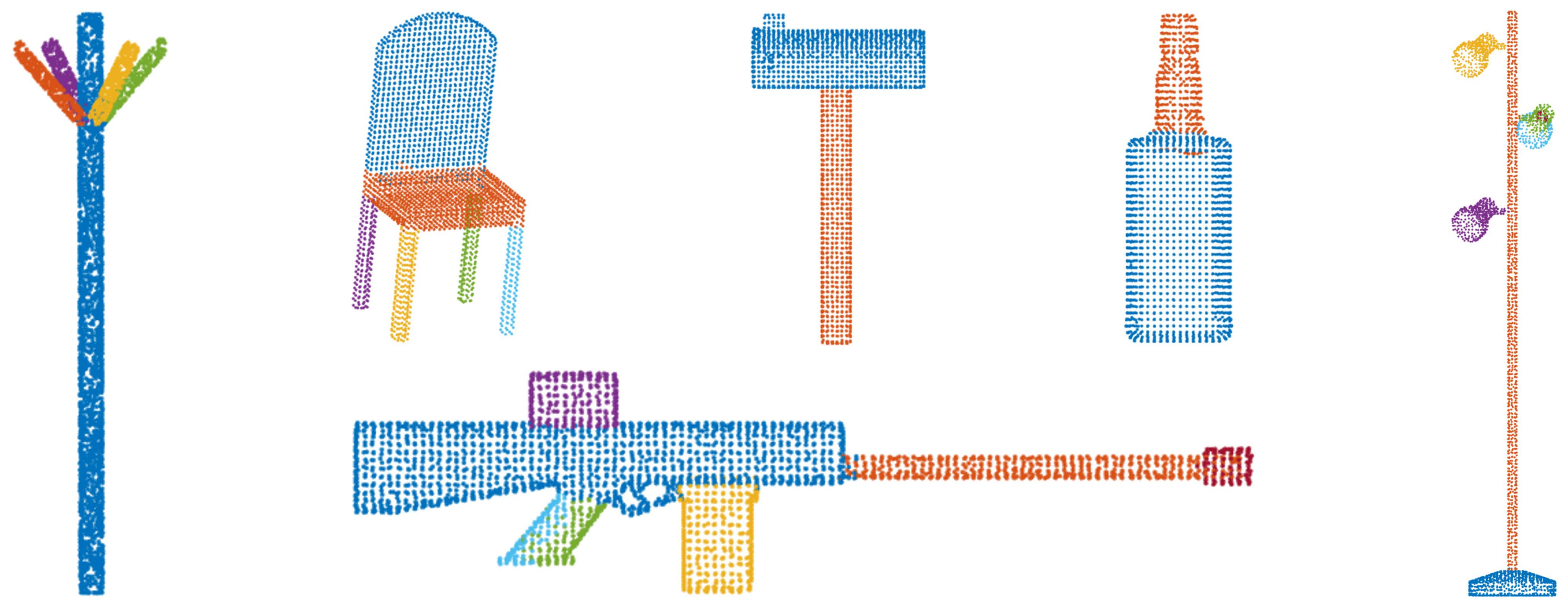}
    \caption{Examples of point clouds segmentation inferred by our method.}
    \label{fig:segmentation}
\end{figure}

Due to the fact that our method yields a geometrically accurate structure, we can achieve a geometry-driven point clouds segmentation task naturally. 
All points in the point clouds have been well clustered after we obtain the abstraction of an object and we segment the point clouds accordingly. 
Fig. \ref{fig:segmentation} illustrates some examples of point clouds segmentation on different objects, inferred by our method.

\section{Conclusions \& Limitations}
In this paper, we present a novel method to abstract the semantic structures of an object.
We cast the problem into a nonparametric clustering framework and solve it by the proposed optimization-based Gibbs sampling.
Additionally, since our method yields a semantically meaningful structure, we can achieve a geometry-driven point clouds segmentation task naturally. 
However, there are some limitations to our method. Firstly, compared with deep learning methods, our implementation is less efficient and cannot be applied in real-time at this moment. In addition, for some
certain categories of objects, such as watercraft and airplanes, which barely consist
of superquadric-like parts, the performance of our algorithm is expected to drop. Furthermore, learning-based methods can produce results with better semantic consistency than ours.

Future work will focus on extending the expressiveness of superquadrics by applying more deformations beyond tapering, such as bending and sheering. 
Additionally, our formulation of how a random point is sampled from a tapered superquadric primitive can be extended to a more general surface beyond superquadrics. 
Moreover, trying different priors for both $\boldsymbol{\theta}$ and $\sigma^2$ other than uniform distributions is also an auspicious way to improve performance.

\vspace{\baselineskip}
\noindent \textbf{Acknowledgments}
This research is supported by the National Research Foundation, Singapore, under its Medium Sized Centre Programme - Centre for Advanced Robotics Technology Innovation (CARTIN) R-261-521-002-592.

\clearpage
%
%
\bibliographystyle{splncs04}
\bibliography{egbib}
\end{document}